\documentclass[letterpaper]{article} 
\usepackage{aaai24}  
\usepackage{times}  
\usepackage{helvet}  
\usepackage{courier}  
\usepackage[hyphens]{url}  
\usepackage{graphicx} 
\usepackage{natbib}  
\usepackage{caption} 
\usepackage{algorithm}
\usepackage{algorithmic}
\usepackage{booktabs}
\usepackage{amsmath}
\usepackage{amssymb}
\usepackage{mathtools}
\usepackage{amsthm}
\usepackage{soul}
\usepackage{multirow}
\usepackage{amssymb}
\usepackage{pifont}
\usepackage{resizegather}
\usepackage[capitalize,noabbrev]{cleveref}

\newtheorem{definition}{Definition}

\newcommand{\indep}{\perp \!\!\! \perp}

\def \bfE {\mathbb{E}}
\def \bfR {\mathbb{R}}

\usepackage{newfloat}
\usepackage{listings}
\DeclareCaptionStyle{ruled}{labelfont=normalfont,labelsep=colon,strut=off} 
\floatstyle{ruled}
\newfloat{listing}{tb}{lst}{}
\floatname{listing}{Listing}
\title{Contrastive Balancing Representation Learning for Heterogeneous \\ Dose-Response Curves Estimation}
\author{
    Minqin Zhu\textsuperscript{\rm 1}\equalcontrib, Anpeng Wu\textsuperscript{\rm 1, 2}\equalcontrib, Haoxuan Li\textsuperscript{\rm 3}, Ruoxuan Xiong\textsuperscript{\rm 4}, Bo Li\textsuperscript{\rm 5}, Xiaoqing Yang\textsuperscript{\rm 6}, Xuan Qin\textsuperscript{\rm 6}, Peng Zhen\textsuperscript{\rm 6}, Jiecheng Guo\textsuperscript{\rm 6}, Fei Wu\textsuperscript{\rm 1}, Kun Kuang\textsuperscript{\rm 1}\thanks{Corresponding author.}
}
\affiliations{
    \textsuperscript{\rm 1} Department of Computer Science and Technology, Zhejiang University\\
    \textsuperscript{\rm 2}Mohamed bin Zayed University of Artificial Intelligence\\
    \textsuperscript{\rm 3}Center for Data Science, Peking University\\
    \textsuperscript{\rm 4}Department of Quantitative Theory and Methods, Emory University\\
    \textsuperscript{\rm 5} School of Economics and Management, Tsinghua University\\
    \textsuperscript{\rm 6}Didi Chuxing\\
    \{minqinzhu, anpwu, kunkuang\}@zju.edu.cn, hxli@stu.pku.edu.cn, ruoxuan.xiong@emory.edu, \\
    libo@sem.tsinghua.edu.cn, \{yangxiaoqing, qinxuan, zhenpeng, jasonguo\}@didiglobal.com, wufei@cs.zju.edu.cn
}

\usepackage{bibentry}

\begin{document}

\maketitle

\begin{abstract}
Estimating the individuals' potential response to varying treatment doses is crucial for decision-making in areas such as precision medicine and management science. Most recent studies predict counterfactual outcomes by learning a covariate representation that is independent of the treatment variable. However, such independence constraints neglect much of the covariate information that is useful for counterfactual prediction, especially when the treatment variables are continuous. To tackle the above issue, in this paper, we first theoretically demonstrate the importance of the \emph{balancing} and \emph{prognostic} representations for unbiased estimation of the heterogeneous dose-response curves, that is, the learned representations are constrained to satisfy the conditional independence between the covariates and both of the treatment variables and the potential responses. Based on this, we propose a novel \textbf{C}ontrastive balancing \textbf{R}epresentation learning \textbf{Net}work using a partial distance measure, called \textbf{CRNet}, for estimating the heterogeneous dose-response curves without losing the continuity of treatments. Extensive experiments are conducted on synthetic and real-world datasets demonstrating that our proposal significantly outperforms previous methods.
\end{abstract}

\section{Introduction}
Causal inference is crucial for individual decision-making, particularly in answering counterfactual questions such as "What would the individual's potential response have been had the person received a different dose of treatment" \cite{raita2021leveraging}. For example, precision medicine is developed by studying the response of drug doses (i.e., continuous treatment) to the potential health state (i.e., potential outcome) of patients (with various medical history information, i.e., covariates) (Shi et al. \citeyear{shi2020selective}). With accessible observational data, an essential obstacle for the unbiased estimation of causal effects is the \emph{confounding bias} from the confounders (i.e., common causes of treatment and outcome), which can lead to spurious correlations between treatment and outcome (Mealli et al. \citeyear{mealli2011statistical}; Pearl et al. \citeyear{pearl2009causality}). Another challenge is the \emph{heterogeneity} of dose-response curves, that is, individuals with different covariates will have different responses even with the same dose given \cite{wager2018estimation,schwab2020learning,wu2023stable,li2023should,li2023trustworthy}.

\begin{table}[tbp]
    \centering
  \resizebox{\linewidth}{!}{
    \begin{tabular}{ll}
    \toprule
Representation Learning Method & Formulation \\
\midrule
Treatment-Balanced  & ${T} \indep \Phi({X})$ \\
Balancing Representation & {${T} \indep {X} \mid \Phi({X})$}\\
Prognostic Representation  & $Y({t}) \indep {X} \mid \Phi({X})$\\
Double Balancing (ours) & $({T}, Y) \indep {X} \mid \Phi({X})$ \\  \bottomrule
\end{tabular}
}
        \caption{A comparison of the constraints employed in various representation learning methods.}
\label{tab:methods_comparing}
\end{table}

Compared with the binary treatment case, dose-response curves has greater challenges in adjusting for the confounding bias of high-dimensional covariates on the continuous treatment (Imai et al. \citeyear{imai2004causal}; Hirano et al. \citeyear{hirano2004propensity}; Kennedy et al. \citeyear{kennedy2017non}). To tackle this problem, the generalized propensity score (GPS) serves as a generalization of the propensity score \cite{rosenbaum1983central} in the binary treatment case, using a Gaussian distribution to model the treatment conditional density for given covariates (Imai et al. \citeyear{imai2004causal}; Hirano et al. \citeyear{hirano2004propensity}). Motivated by covariate balancing propensity score \cite{imai2014covariate, hainmueller2012entropy}, the optimal balancing weighting methods focus on learning sample weights such that the treatment and covariates are independent on the re-weighted data \cite{fong2018covariate,vegetabile2021nonparametric}. Despite focusing on the \emph{unbiasedness} that treatments and confounders are conditionally independent on balancing scores, these methods show limited performance in practice when the covariates are high-dimensional \cite{nie2021vcnet, schwab2020learning}. Furthermore, these methods neglect the outcome during the modeling process of the balancing weight, which might omit important confounders that are necessary for outcome prediction \cite{stuart2013prognostic, hansen2008prognostic}.

With the progress of deep learning, recent studies apply neural networks to fit dose-response curve of high-dimensional covariates (Bica et al. \citeyear{bica2020estimating};  Schwab et al. \citeyear{schwab2020learning}; Nie et al. \citeyear{nie2021vcnet}). In deep methods, a critical challenge is how to learn appropriate covariate representations for heterogeneous dose-response curve estimation \cite{kallus2020deepmatch}. 
Specifically, DRNet \cite{schwab2020learning} propose to learn treatment-balanced representations (Shalit et al.  \citeyear{shalit2017estimating}; Wu et al. \citeyear{wu2022instrumental}; Wang et al. \citeyear{wang2023optimal}) which force the learned representations to be independent of continuous treatments. Nonetheless, the method's \emph{unbiasedness} hinges on the assumption of invertibility concerning covariate representations, which is stringent for deep methods \cite{behrmann2019invertible}. 
In practice, imposing the constraint of independence between treatment assignment and covariate representations runs the risk of neglecting confounder information that is essential for outcome prediction, leading to biased estimates \cite{assaad2021counterfactual}. To tackle this problem, VCNet \cite{nie2021vcnet} employs a propensity score estimator to constrain the representations for unbiased average dose-response curve estimation. However, while the propensity score is the coarsest balancing score \cite{rosenbaum1983central}, it may not adaptable for the heterogeneous dose-response curve estimation because of the covariate information loss \cite{hahn1998role}. In other words, it might not \emph{prognostic} that potential outcomes and confounders are conditionally independent given balancing scores \cite{hansen2008prognostic}. For the prognostic representation, SCIGAN \cite{bica2020estimating} directly models the treatment effect by generative adversarial networks \cite{goodfellow2020generative}. However, it is noteworthy that SCIGAN does not explicitly account for the unbiasedness.

Overall, obtaining appropriate representations that eliminates confounder bias and retains necessary confounder information for the unbiased heterogeneous dose-response curve is still a challenging problem~\cite{wu2022learning}.
To solve this problem, we systematically introduce the double balancing representation, i.e., a combination of the balancing and prognostic representations, which is constrained to satisfy the conditional independence between the covariates and both of the treatments and the potential outcomes. 
For the double balancing representation, we propose a novel contrastive regularizer, applying contrastive learning \cite{chen2020simple, he2020momentum, grill2020bootstrap} to monitor the unbiasedness condition and maintain treatment continuity. 
Specifically, we create negative samples by randomly shuffling the original covariates and treat the original covariates as positive samples \cite{arbour2021permutation, cheng2020club}. Adaptable to the cross-entropy loss \cite{chen2020simple}, we adopt partial distance measure \cite{szekely2014partial} to evaluate the unbiasedness condition and design a contrastive regularizer loss to minimize the partial distance measure while discriminating among positive and negative samples. Moreover, to preserve the predictive power of the representation for the outcome, we design a mean squared error loss specifically tailored to address prognostic representation. Empirically, we demonstrate that {CRNet} achieves state-of-the-art performance on both synthetic and semi-synthetic datasets with different dimensions of continuous treatments. We summarize our contribution as follows:
\begin{itemize}
 \item[$\bullet$] For unbiased heterogeneous dose-response curve estimation, we systematically define a double balancing representation condition which satisfies the conditional independence constraint between the covariates and both of the continuous treatments and the observed outcomes.

  \item[$\bullet$] We propose a novel {CRNet} architecture for learning double balancing representations without losing the continuity of treatments. Specifically, we design a contrastive loss with a partial distance measure of positive and negative samples and a mean square error loss to optimize the CRNet. To the best of our knowledge, this is the first paper to apply contrastive learning in the field of heterogeneous dose-response curve estimation.

\item[$\bullet$] Empirically, varying the dimension of continuous treatments and covariates in both simulated and real-world datasets, we demonstrate that the proposed {CRNet} outperforms other baseline methods on HDRC estimation.
\end{itemize}

\section{Related Work}
\paragraph{Dose-Response Curve Estimation.}
For estimating the dose-response curve\footnote{We only discuss weighting methods because matching and stratification can be considered as particular forms of weighting.}, traditional methods (Imbens et al. \citeyear{imbens2000role}; Imai et al. \citeyear{imai2004causal}; Fong et al. \citeyear{fong2018covariate}; Vegetabile et al. \citeyear{vegetabile2021nonparametric}) learn sample weights on selected metrics to achieve the balance of covariates to eliminate the confounding bias. However, these methods neglect the outcome during the modeling of the balancing weight, which might omit confounders that are necessary for outcome prediction \cite{hansen2008prognostic,stuart2013prognostic,lee2022review}.

Deep methods learn appropriate representations for DRC 
 estimation (Bica et al. \citeyear{bica2020estimating};  Schwab et al. \citeyear{schwab2020learning}; Nie et al. \citeyear{nie2021vcnet}). Treatment-balanced representation methods, for instance, DRNet \cite{schwab2020learning} constrains representations independent of continuous treatments. VCNet \cite{nie2021vcnet} and SCIGAN \cite{bica2020estimating} constrain representations by treatment estimators/discriminators. None of them explicitly constrain that the learned representation satisfies both balancing and prognostic representation conditions for dose-response curve estimation.
Instead, we propose a novel contrastive regularizer network to obtain double balancing representations for unbiased heterogeneous dose-response curve estimation directly.

\paragraph{Contrastive Representation Learning.} 
Contrastive representation learning \cite{chen2020simple, he2020momentum, grill2020bootstrap, DBLP:conf/sigir/ZhangYZC021, zhang2022tree,DBLP:conf/www/YaoZZZZZH22,gan-etal-2023-exploiting} is a self-supervised learning method. It approximates the latent representations by constructing contrastive samples (positive and negative instances) to facilitate instance discrimination \cite{wu2018unsupervised}. 
Through the process of discriminating between contrastive samples, positive instances are closer to the original instance in the projection space, while negative instances are further away from the original instance in the projection space to maximize the lower bound of the mutual information \cite{wang2020understanding, huang2021towards}. 
In this paper, we apply contrastive learning to regularize this representation without breaking the continuity of treatments. To the best of our knowledge, this is the first paper to apply contrastive learning in heterogeneous dose-response curve estimation.

\section{Problem Setup}

For the case of continuous treatments, we observe $n$ units with baseline covariates $X \in \mathcal{X}\subset \mathbb{R}^p$, continuous treatments $T \in \mathcal{T}\subset \mathbb{R}^q$ and outcome $Y \in \mathcal{Y}\subset \mathbb{R}$, where $p,q$ is dimension of covariates and treatments, respectively.  We also let $\mathbf{X}\in \mathcal{X}^n \subset \mathbb{R}^{n\times p}$, $\mathbf{T}\in \mathcal{T}^n \subset \mathbb{R}^{n\times q}$ and $\mathbf{Y}\in \mathcal{Y}^n \subset \mathbb{R}^n$ denote all the observed baseline covariates,
continuous treatments, and outcomes, respectively. Using Neyman-Rubin potential
outcome framework \cite{rubin1974estimating,rosenbaum1983central}, for an observation for unit $i$ with received $T_i = t$, there is a potential outcome $Y_i(t)$. 

Throughout this paper, we assume three assumptions that are commonly made in continuous treatment settings \cite{imbens2000role,schwab2020learning,nie2021vcnet,bica2020estimating}. Specifically, for a unit $i$, we assume the stable unit treatment value assumption (SUTVA) assumption holds that we can only observe the potential outcome corresponding to the received treatment level $t$, i.e., $Y_i = Y_i(t)$ and there should not be alternative forms of the treatment and interference between units, capturing consistency and non-interference. Moreover, we assume the unconfoundedness assumption that $Y(t) \indep T \mid X$ and the positivity assumption that $0 < \mathbb{P}(t|x)$ for $T=t$ and $X=x$. In this paper, $\indep$ denotes (conditional) independence, and $\mathbb{P}$ is the probability density function (pdf).
We consider estimating the heterogeneous dose-response curve (HDRC):
\begin{equation}
h(t,x) = \mathbb{E}[Y(t) \mid X=x],
\end{equation}
where $\bfE$ denotes expectation.

\section{Motivation}
For estimating the heterogeneous dose-response curve,
deep methods require an appropriate criterion to monitor the representation they produce \cite{kallus2020deepmatch, schwab2020learning}. Inspired by the effective balancing score \cite{hu2014estimation, huang2017joint}, a linear function of covariates for unbiased causal effect estimation, we turn to define two conditions of representation for unbiased heterogeneous dose-response curve estimation.
\begin{definition}[Balancing Representation Condition]
\label{def:BR}
A balancing representation $\Phi({X}), {X} \in \mathcal{X}$, correlated to treatments ${T}\in\mathcal{T}$ and outcome $Y \in \mathcal{Y}$ satisfies: 
\begin{equation}
\begin{aligned}\label{balancing_representation}
{X} \indep {T} \mid \Phi({X}).
\end{aligned}
\end{equation} 
\end{definition}
Theoretically, let $F_X(\cdot|\cdot)$ denote conditional probability distributions for $X$, we can derive the treatment assignment: 
\begin{equation}
\begin{aligned}
&\mathbb{P}_T(T = t|\Phi(x),Y(t)) \\
& = \int_{x^\prime} \mathbb{P}_T(T=t|X={x^\prime} ,\Phi(x),Y(t))dF_X(X={x^\prime} |\Phi(x)
, Y(t))
\\
&=\int_{x^\prime}  \mathbb{P}_T(T=t|X={x^\prime} )dF_X(X={x^\prime} |\Phi(x))=\mathbb{P}_T(t|\Phi(x)).
\end{aligned}
\end{equation}

The first and third equations hold by using iterated expectation operation, the second equation holds by unconfoundedness assumption and \cref{def:BR}. The above equation implies that $X\indep T\mid \Phi(X)$ is equivalent to the unbiasedness condition, i.e., $Y(t)\indep T\mid \Phi(X)$. It guarantees that the treatment assignment is ignorable given the balancing representation when the unconfoundedness assumption is satisfied \cite{rosenbaum1983central}.
As a result, we can identify the average dose-response curve as:
\begin{equation}
\begin{aligned}
\bfE[Y(t)]&=\bfE_X[\bfE[Y(t)\mid \Phi(x)]]\\
&=\bfE_X[\bfE[Y(t)\mid \Phi(x), T=t]]\\
&=\bfE_X[\bfE[Y\mid\Phi(x), T=t]].   
\end{aligned}
\end{equation}

The first equation holds by the iterated expectation, the second equation holds by $Y(t)\indep T\mid \Phi(X)$, and the third equation holds by the consistency assumption in SUTVA. 
Nevertheless, regressing outcome $Y$ on balancing representation $\Phi(X)$ and treatment $T$ is inadequate for the heterogeneous dose-response curve estimation. This inadequacy can be attributed to the following reasons:
\begin{equation}\label{proof:br}
    \bfE[Y(t)\mid x]\neq \bfE[Y(t)\mid \Phi(x)].
\end{equation}

For instance, VCNet \cite{nie2021vcnet}, which employs a propensity score constraint on representation, is sufficient for unbiased average dose-response estimation. Nonetheless, it is important to acknowledge that the propensity score is the coarsest balancing score \cite{rosenbaum1983central}. Using representations that are constrained by it might result in the loss of covariate information for outcome prediction \cite{hahn1998role}. This concern is amplified in situations where VCNet discretizes the continuous treatment variable into discrete intervals and utilizes the cross-entropy loss function for training the propensity score \cite{li2023propensity}. 


\begin{definition}[Prognostic Representation Condition]
\label{def:PR}
A prognostic representation $\Phi({X}), {X} \in \mathcal{X}$ correlated to treatments ${T}\in\mathcal{T}$ and outcome $Y \in \mathcal{Y}$ satisfies: 
\begin{equation}\label{eq:PR}
\begin{aligned}
{X} \indep {Y}(t) \mid \Phi({X}).
\end{aligned}
\end{equation} 
\end{definition}

Theoretically, the prognostic representation condition in Eq. (\ref{eq:PR}) is sufficient for unbiased heterogeneous dose-response curve estimation. Consider a unit $i$ with treatment $t$, we can write the heterogeneous dose-response curve as:
\begin{equation}
\begin{aligned}
\resizebox{.9\hsize}{!}{$
\bfE[Y(t)\mid x]=\bfE[Y(t)\mid \Phi(x)] = \bfE[Y\mid \Phi(x), T=t].
$}
\end{aligned}
\end{equation}

The first equation holds by \cref{def:PR} and the second equation holds because the prognostic representation is also a balancing representation \cite{hansen2008prognostic,stuart2013prognostic}. The above analysis implies that given the prognostic representation, the covariates are ignorable for the outcome prediction \cite{hansen2008prognostic}. 

Learning the prognostic representation presents a challenge due to the unobservability of potential outcomes $Y(t)$ \cite{holland1986statistics}.  In practical scenarios, we are constrained to derive representations based on the condition $ {X} \indep {Y} \mid \Phi({X})$, which we refer to as the \emph{learnable} prognostic representation condition, a condition also implicitly utilized in SCIGAN (Bica et al. \citeyear{bica2020estimating}). Consequently, when representations are constrained by this prognostic representation condition between covariates and \emph{observed} outcomes,  there exists a potential concern regarding ensuring unbiasedness \cite{hansen2008prognostic,huang2017joint}.


To address the challenges posed by both the balancing representation and prognostic representation conditions, where balancing representation condition alone may lead to the loss of essential information for outcome prediction and prognostic representation condition constrained by observed outcomes may introduce bias, we propose the condition of double balancing representation. This condition aims to enhance both unbiasedness through the balancing representation condition and the predictive capacity for outcomes through the prognostic representation condition.

\begin{definition}[Double Balancing Representation Condition]
\label{def:DBR}
A double balancing representation $\Phi({X}), {X} \in \mathcal{X}$ correlated to treatments ${T}\in\mathcal{T}$ and outcome $Y \in \mathcal{Y}$ satisfies: 
\begin{equation}
\begin{aligned}
{X} \indep {T} \mid \Phi({X}), \; \; {X} \indep {Y} \mid \Phi({X}).
\end{aligned}
\end{equation}  
\end{definition}

Theoretically, consider a unit $i$ with treatment $t$. Given the double balancing representation, we can identify the heterogeneous dose-response curve as:
\begin{equation}\label{proof:eqr}
\begin{aligned}
\mathbb{E}[Y(t)|x]=\mathbb{E}[Y|x,T=t]=\mathbb{E}[Y|\Phi(x),T=t].\\
\end{aligned}
\end{equation}
It becomes evident that the second equation is valid when the condition of double balancing representation is met. In the context of HDRC estimation, the two conditions comprising the double balancing representation are of paramount importance, as they mutually reinforce the effectiveness of this representation. Hence, it is imperative to verify the fulfillment of the double balancing representation condition.

\section{Method}
The two conditions in double balancing representation both control confounder information of treatment assignment and retain necessary confounder information for outcome prediction. For unbiased heterogeneous dose-response curve estimation, we design the contrastive regularizer loss and mean square loss to constrain these two conditions.

\subsection{Contrastive Regularizer}
In the setting of dose-response curve estimation, the treatments can be multiple and continuous and the covariates are high-dimensional.
To ensure the unbiased treatment assignment, it is necessary to quantify the conditional dependence of $T$ and $X$ given $\Phi(X)$ \cite{rosenbaum1983central}. Without loss of generalization,
we adopt partial distance measure \cite{szekely2014partial} to achieve this goal. The partial distance measure is a scalar quantity that captures dependence, which equals the conditional correlation in Gaussian scenarios. In the non-Gaussian case, a partial distance of zero does not confirm conditional independence, however, such a measure that is closer to zero indicates a weaker association (refer to Sec 4.2. in \cite{szekely2014partial}).

\paragraph{Partial Distance Measure.} For all observed data, assuming three variables $\mathbf{X},\mathbf{T},
\boldsymbol{Z}$ and their double-centered pairwise distance $\omega(\mathbf{X})$, $\omega(\mathbf{T})$, $\omega(\boldsymbol{Z})$, we define $[\omega(\mathbf{X})]_{i, j}=$
$\left\|\mathbf{X}_{i}-\mathbf{X}_{j}\right\|
     -\frac{1}{n}\sum_{k=1}^n \left\|\mathbf{X}_{k}
     -\mathbf{X}_{j}\right\|-\frac{1}{n}$
$\sum_{l=1}^n \left\|\mathbf{X}_{i}-\mathbf{X}_{l}\right\|$ 
$+\frac{1}{n^2}\sum_{k=1}^n\sum_{l=1}^n\left\|\mathbf{X}_{k}-\mathbf{X}_{l}\right\|$
where
$\left\|\cdot\right\|$ is the Euclidean norm and $\omega(\cdot) \in \bfR^{n\times n}$.
The form of $\omega(\mathbf{T}), \omega(\boldsymbol{Z})$ are similar. 
We define the inner product of $\omega(\mathbf{X})$ and $\omega(\mathbf{T})$ as $\omega(\mathbf{X}) \otimes \omega(\mathbf{T})=[{n(n-3)}]^{-1} \sum_{i \neq j} [\omega(\mathbf{X})]_{i, j} \cdot[\omega(\mathbf{T})]_{i, j}.$

The double-centered pairwise distance orthogonal projection of $\mathbf{X}$ on $\boldsymbol{Z}$ is $\operatorname{proj}_{\boldsymbol{Z}}(\mathbf{X})=\omega(\mathbf{X})-\omega(\mathbf{X}) \otimes \omega(\boldsymbol{Z})$ $[{\omega(\boldsymbol{Z}) \otimes \omega(\boldsymbol{Z})}]^{-1}\omega(\boldsymbol{Z}),$
and the projection of $\omega(\mathbf{T})$ on $\omega(\boldsymbol{Z})$ is similar.
Then, we formulate the partial distance measure $D_{\boldsymbol{Z}}(\mathbf{X}, \mathbf{T})$ as follows:
\begin{equation}
    D_{\boldsymbol{Z}}(\mathbf{X}, \mathbf{T})= \frac{\left|\operatorname{proj}_{\boldsymbol{Z}}(\mathbf{X}) \otimes \operatorname{proj}_{\boldsymbol{Z}}(\mathbf{T})\right|}{\left\|\operatorname{proj}_{\boldsymbol{Z}}(\mathbf{X})\right\|\cdot\left\|\operatorname{proj}_{\boldsymbol{Z}}(\mathbf{T})\right\|},
\label{positive_loss}
\end{equation}
where $|\cdot|$ is the absolute operation. The norm $\left\|\operatorname{proj}_{\boldsymbol{Z}}(\mathbf{X})\right\|$ $=(\operatorname{proj}_{\boldsymbol{Z}}(\mathbf{X}) \otimes \operatorname{proj}_{\boldsymbol{Z}}(\mathbf{X}))^{1/2}$ and the norm $\left\|\operatorname{proj}_{\boldsymbol{Z}}(\mathbf{T})\right\|$ similarly defined. 

Given the partial distance measure, a key challenge relates to the design of the loss function \cite{lecun2006tutorial}. The motivation for employing contrastive learning stems from the potential mode collapse issue 
\cite{jing2021understanding,goodfellow2020generative}, which can occur when naively minimizing partial distance measures for positive samples. Mode collapse is a fundamental problem in representation learning \cite{he2020momentum,chen2020simple,chen2021exploring}and arises when a model fails to adequately capture the diverse patterns within the data, instead collapsing them into a single mode or a limited set of modes. For instance, if the balancing representation exhibits a multi-modal distribution, but the model only learns a uni-modal distribution, it becomes susceptible to mode collapse. This situation can introduce bias into the learned representation for the estimation of heterogeneous dose-response curves \cite{li2023propensity}. To obviate it, we adopt the contrastive learning \cite{chen2020simple}.

\begin{figure*}[ht]
  \centering
  \includegraphics[width=0.9\linewidth]{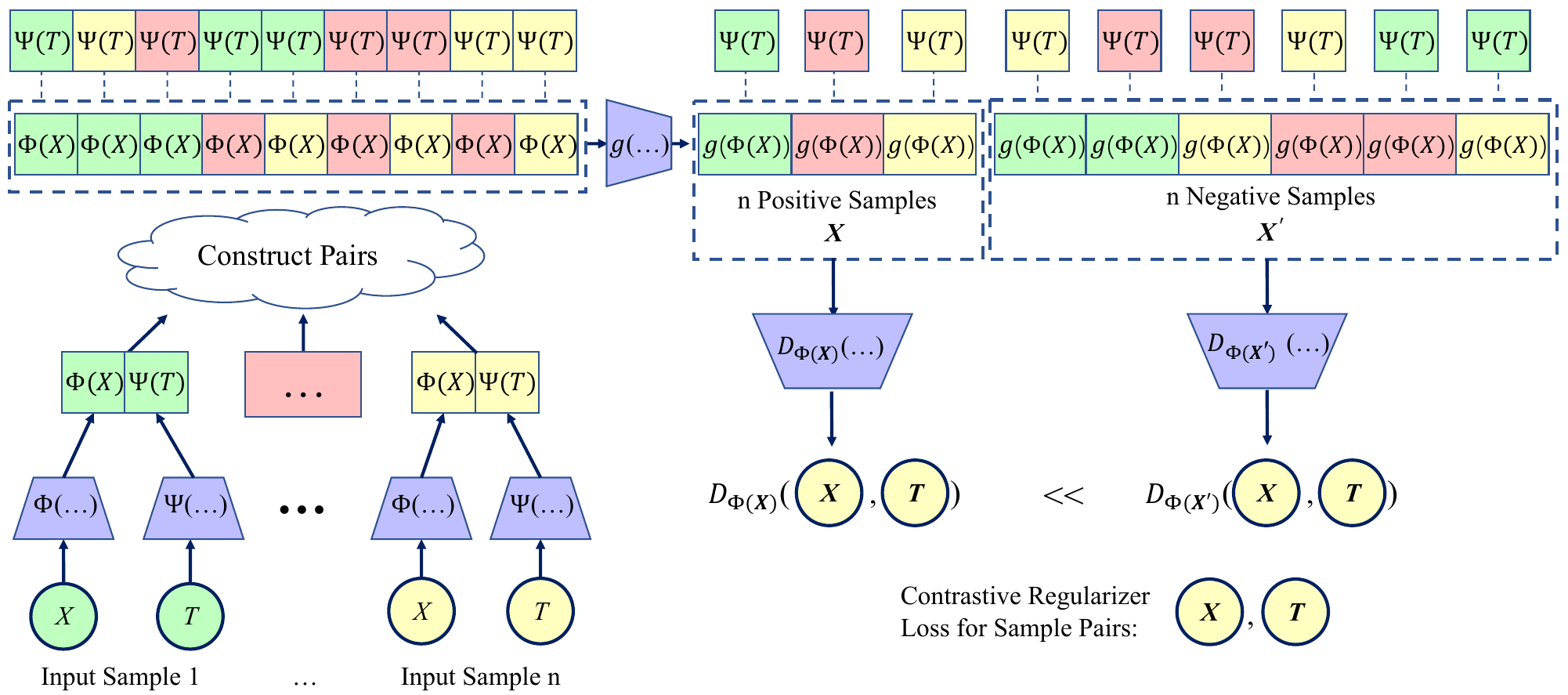}
  \caption{Contrastive regularizer. The $n$ covariates $\mathbf{X}$ undergo a transformation via the encoder $\Phi$ resulting in $\Phi(\mathbf{X})$. $\Phi(\mathbf{X})$ transforms to $g(\Phi({\mathbf{X}}))$ through projection head $g$ \cite{chen2020simple}. $g(\Phi({\mathbf{X}}))$ is directly constrained by $\ell^{CR}_{\Phi}(\mathbf{X}, \mathbf{T})$. To simplify notation, we use $\Phi(\mathbf{X})$ in the context to represent $g({\Phi(\mathbf{X}}))$. $D_{\Phi({\mathbf{X}})}$ and $D_{\Phi({\mathbf{X}'})}$ are partial distance measure of positive/negative samples \cite{szekely2014partial}.}
  \label{contrastive_loss}
\end{figure*}

We propose a novel contrastive regularizer (CR) to constrain the balancing representation. To construct the positive and negative samples for contrastive learning, we randomly shuffle the $n$ units of $\mathbf{X}$ in the original data $m$ times to get $n*m$ permuted data $\mathbf{X}'$ that $\mathbf{X}'\perp\!\!\!\perp \mathbf{T}$ \cite{arbour2021permutation, cheng2020club}. Then we name $\mathbf{X}$ \emph{positive samples} and $\mathbf{X}'$ \emph{negative samples} and define their partial distance measure as:
\begin{equation}
\resizebox{.8\hsize}{!}{$
{D}_{\Phi(\mathbf{X})}(\mathbf{X}, \mathbf{T})=\frac{\left|\operatorname{proj}_{\Phi(\mathbf{X})}(\mathbf{X}) \cdot \operatorname{proj}_{\Phi(\mathbf{X})}(\mathbf{T})\right|}{\left\|\operatorname{proj}_{\Phi(\mathbf{X})}(\mathbf{X})\right\|\cdot \left\|\operatorname{proj}_{\Phi(\mathbf{X})}(\mathbf{T})\right\|}, 
$}
\end{equation}

\begin{equation}
\resizebox{.8\hsize}{!}{$
    {D}_{\Phi(\mathbf{X}')}(\mathbf{X}, \mathbf{T})=\frac{\left|\operatorname{proj}_{\Phi(\mathbf{X}')}(\mathbf{X}) \cdot \operatorname{proj}_{\Phi(\mathbf{X}')}(\mathbf{T})\right|}{\left\|\operatorname{proj}_{\Phi(\mathbf{X}')}(\mathbf{X})\right\|\cdot \left\|\operatorname{proj}_{\Phi(\mathbf{X}')}(\mathbf{T})\right\|}.
$}
\end{equation}

As the Fig. \ref{contrastive_loss} shown, given the representations $\Phi(\mathbf{X})$ and $\Phi(\mathbf{X}')$ from $n$ observed covariates $\mathbf{X}$ and $n$ shuffled covariates $\mathbf{X}'$, the correctly specified function $\Phi$ should satisfy that ${D}_{\Phi(\mathbf{X})}\left(\mathbf{X}, \mathbf{T}\right) \ll {D}_{\Phi(\mathbf{X}')}\left(\mathbf{X}, \mathbf{T}\right)$. Then we propose to perform contrastive learning for the positive samples $\mathbf{X}$ and negative samples $\mathbf{X}'$. The contrastive regularizer loss is formulated as follows: 
\begin{equation}\label{CRloss}
\begin{aligned}
\resizebox{.95\hsize}{!}{$
\ell^{CR}_{\Phi}(\mathbf{X}, \mathbf{T})=D_{\Phi(\mathbf{X})}(\mathbf{X},\mathbf{T})
-\log {\sum_{j=1}^{m} \exp{\left({D}_{\Phi(\mathbf{X}'_{(j)})}(\mathbf{X}, \mathbf{T})\right)}},
$}
\end{aligned}
\end{equation}
where $m$ represents the number of shuffles, and $\mathbf{X}'_{(j)}$ denotes the shuffled covariates from the $n$ negative samples acquired during the $j$th shuffle. The total count of negative samples is given by $n \times m$.
During the training procedure, for each batch of samples, we perform random shuffling of the original covariates within the batch a total of $m$ times, and we set $m=1$ with default \cite{cheng2020club}.

It is worth noting that the contrastive regularizer serves a dual purpose, not only preserving unbiasedness but also ensuring the continuity of treatments, thereby benefiting prognostic condition, owing to its enhancement of representation quality. More specifically, given that all discrimination operates at the instance level \cite{wu2018unsupervised}, there is no necessity to discretize treatment variables into bins \cite{schwab2020learning,bica2020estimating,nie2021vcnet}. Consequently, the continuity of treatments is inherently maintained. Furthermore, the contrastive regularizer with positive/negative samples effectively captures diverse information from covariates $X$ for representation learning. This aspect aligns with the requirement of addressing prognostic condition, aiming to capture the differences in causal effects among various study subjects \cite{hansen2008prognostic}.

\subsection{CRNet}
Different from the unbiasedness condition, which focuses on the treatment assignment. The prognostic condition focus on the outcome prediction power in representation. In this paper, we design a two-head neural network, which encodes the treatments $T$ through $\Psi$ and covariates $X$ through $\Phi$ for representations $\Psi(T)$ and $\Phi(X)$. Then, we adopt a mean square error loss (MSE) to directly constrain the condition in the double balancing representation that $Y\indep X|\Phi(X)$. In particular, for a unit $i$, the MSE loss is formulated as follows:
\begin{equation}
    \ell^{{MSE}}({X}_i, {T}_i, Y_i) = (Y_i-h(\Phi({X}_i), \Psi({T}_i)))^2.
\end{equation}

Although the MSE loss has been commonly employed in previous works for outcome prediction \cite{schwab2020learning,bica2020estimating,nie2021vcnet}, it's important to note that most of these approaches do not constrain the prognostic condition effectively. To elaborate, DRNet \cite{schwab2020learning} imposes the MSE loss on a treatment-balanced representation that is independent of treatments $T$. However, as treatments $T$ is correlated with covariates $X$, this approach may lead to a loss of essential confounder information for outcome prediction. VCNet \cite{nie2021vcnet} utilizes the MSE loss on representations constrained by a propensity score estimator, which is considered the coarsest balancing score \cite{rosenbaum1983central}. Representations subject to this constraint may also fail to satisfy the prognostic representation condition \cite{hahn1998role}. Furthermore, these methods tend to neglect the issue of mode collapse, which can compromise their ability to estimate heterogeneous causal effects effectively. 
Different from them, our MSE loss is imposed on double balancing representations, while constraining the representation to satisfy unbiasedness. We achieve this by employing the contrastive regularizer to preserve the confounder information of $X$. This facilitates the direct regression model to learn prognostic representation as much as possible.

\begin{figure}[ht]
  \centering
  \includegraphics[width=8cm]{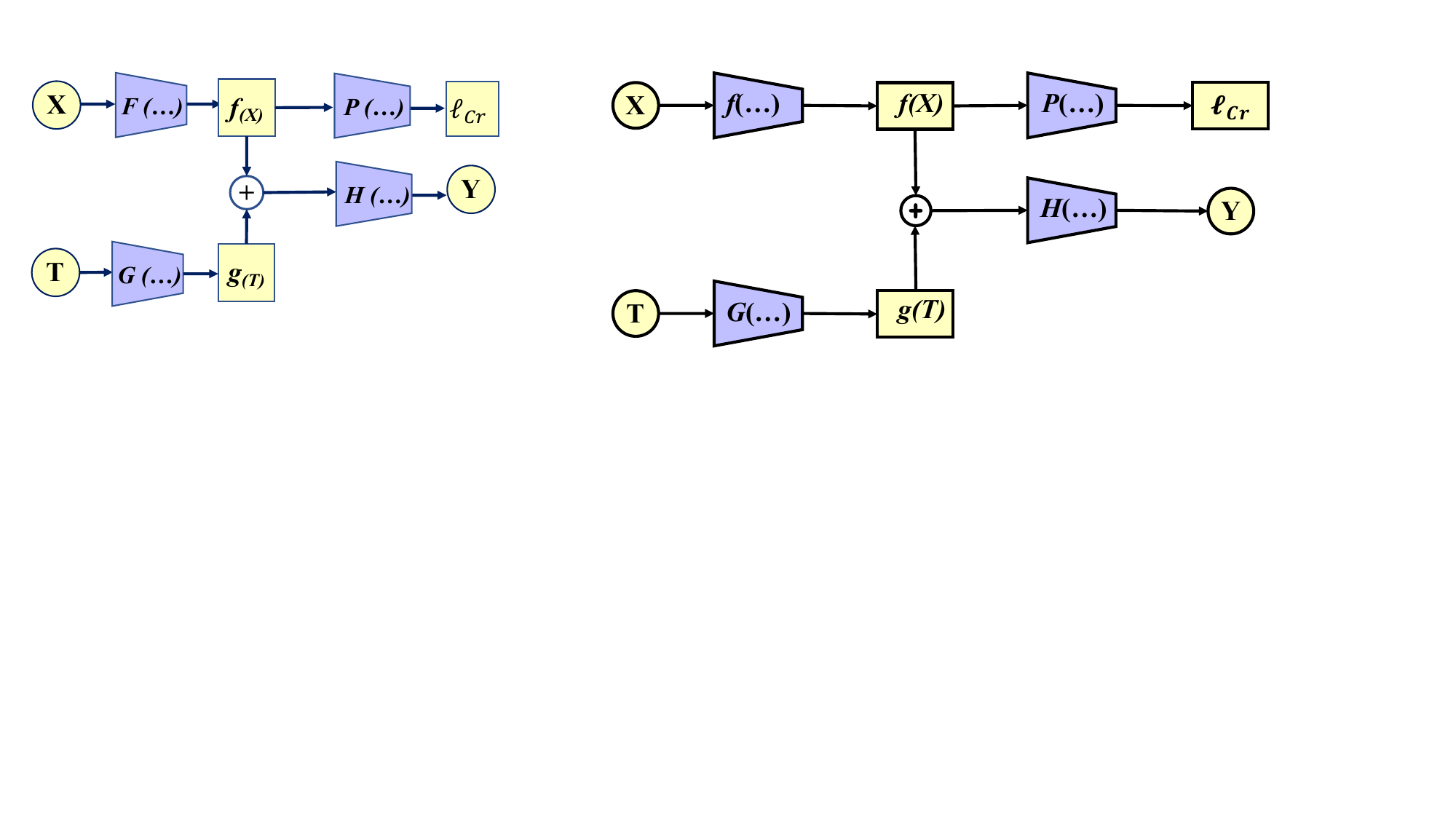}
  \caption{\emph{CRNet}. For the training procedure, the representations $\Phi(\mathbf{X})$ constrained by contrastive loss $\ell^{CR}_{\Phi}(\mathbf{X},\mathbf{T})$ are concatenated and input to MLPs $h$ to obtain the estimated outcome $\hat{Y}$ by the final loss in Eq. (\ref{finalloss}). The final objective is to minimize the loss. For the inference procedure, the estimated HDRC is obtained by $h(\Phi(\mathbf{X}),\Psi(\mathbf{T}))$.}
  \label{CRNet}
\end{figure}

In summary, we propose a neural network framework called CRNet for the estimation of HDRC. As depicted in Figure \ref{CRNet}, the overall architecture of CRNet comprises three distinct blocks: a) Two-Head Encoder. The first head, denoted as $\Phi$, encodes covariates $\mathbf{X}$ into representation $\Phi(\mathbf{X})$. The second head, denoted by $\Psi$, encodes treatments $\mathbf{T}$ into representation $\Psi(\mathbf{T})$;
b) Projection Head. The projection head, denoted by $g$, project the covariate representation $\Phi(\mathbf{X})$ into $g(\Phi(\mathbf{X}))$ for the partial distance measure involving covariates $\mathbf{X}$ and treatments $\mathbf{T}$;
c) Outcome Estimator. The outcome estimator $h$  takes the concatenated representations of covariates and treatments, $\Phi(\mathbf{X})$ and $\Psi(\boldsymbol{T)}$, as input and transforms them into $h(\Phi(\mathbf{X}),\Psi(\mathbf{T}))$. This estimated outcome $h(\Phi(\mathbf{X}),\Psi(\mathbf{T}))$ approximates the observed $Y$ by the regularized regression loss $\ell^{final}(\mathbf{X},\mathbf{T},\mathbf{Y})$:
\begin{equation}
\label{finalloss}
\begin{aligned}
\resizebox{.98\hsize}{!}{$
 \ell^{final}(\mathbf{X},\mathbf{T},\mathbf{Y}) =\sum_{i=1}^{n}\ell^{{MSE}}({X}_i, {T}_i, Y_i)
 + \alpha*\ell^{CR}_\Phi(\mathbf{X},\mathbf{T}),  
$}
\end{aligned}
\end{equation}
where $\alpha$ represents the hyperparameter.
The use of both the MSE loss $\ell^{MSE}$ and the contrastive regularizer loss $\ell^{CR}_{\Phi}$ is necessary for unbiased estimation of HDRC. The loss $\ell^{MSE}$ is used to minimize the difference between the predicted outcome and the observed outcome, thus ensuring that the model can make reliable predictions. Without this loss, the CRNet model would not be able to accurately predict the outcomes, resulting in loss of prediction capacity. On the other hand, the loss $\ell^{CR}_{\Phi}$ helps to prevent treatment assignment bias and outcome overfitting by comparing the representation of positive/negative samples.  Without this loss, the CRNet model might induce treatment assignment bias or mode collapse, leading to inaccurate outcome prediction. In short, the two losses in $\ell^{final}$ complement each other.

\section{Experiments}\label{experiments}

Since the true HDRC are rarely available in real application, in line with previous work (Nie et al. \citeyear{nie2021vcnet}; Bica et al. \citeyear{bica2020estimating}), we simulate 4 synthetic data and 5 semi-synthetic data from two real-world datasets {IHDP}\footnote{https://www.fredjo.com} and {News}\footnote{https://paperdatasets.s3.amazonaws.com/news.db}.

\subsection{Experimental Setup}
\paragraph{Synthetic Data Generation.} We simulate synthetic data as follows. For each unit $i \in \{1,2,\cdots, n\}$, we generate $p=100$ covariates from an independent identical distribution, i.e., $\mathbf{X}_i \sim \mathcal{N}(\mathbf{0}_p,\mathbf{E}_p)$, where $\mathbf{0}_p$ denotes a $p$-dimensional vector with all elements equal 0, and $\mathbf{E}_p$ represents $p$-order identity matrix. 
We generate $q$ treatments using the following rules:
$T_{i,j} = 0.2\sum_{j=1}^5 W_j X_{i,j} + \frac{1}{p-11}\sum_{j=11}^p W_j X_{i,j}^2 
    + \tilde{T}_{i,j} + 0.5\tilde{T}_{i,j}^3 X_{i,p-j}.$ 
Here, we denote $ \tilde{T}_{i,j} \sim$ $ \mathcal{N}(0,1), $ $W_j \sim U(0.5,1)$ for $j \in \{1,2, \cdots, q\}$.
The outcome is generated according to the following rules:
$ Y_i=0.5 \sum_{j=1}^{q}{W}_j^{T}  |{T}_{i,j}| + 0.5\sum_{j=6}^{10} $ $ W_j{X}_{i,j}^2 +
 \sum_{j=11}^{p}{W}{X}_{i} + 0.5 \sum_j^{q} {T}_{i,j}{X}_{i,q-j-10}.  $
Here, we denote $W_j^{T} \sim U(0.5,1)$.
Then, we design 4 simulation datasets and name them Data-$q$ where $q$ means the dimension of $\mathbf{T}$ and $\mathbf{X}$ (e.g., Data-1 means a simulation with 1 treatment, 100 covariates). Then we sample 2100/600/300 units for training/validation/test for each data.

\paragraph{Semi-synthetic Data Generation.} We proceed to perform semi-simulation experiments with the aim of demonstrating the robustness of our method across a range of settings. These settings are designed in accordance with the data generation rules of synthetic data generation. We sample units from the IHDP data to create the training, validation, and test sets, with 522/150/75 units for each data split. 
For the News dataset, we perform data splits into training, validation, and test sets with 2100/600/300 units, respectively.

\paragraph{Baselines and Evaluation.} We compare our model with the following baselines in the above datasets: For statistical methods, we use (1) \textbf{Causal Forest} (Wager et al. \citeyear{wager2018estimation}), a random forest algorithm for causal inference. (2) \textbf{GPS} \cite{imbens2000role}, a generalized propensity score for continuous treatments. (3) \textbf{CBGPS} (Fong et al. \citeyear{fong2018covariate}), a generalized covariate balancing propensity score \cite{imai2014covariate} for continuous treatments. 
For deep methods, we apply (4) \textbf{SCIGAN} (Bica et al. \citeyear{bica2020estimating}), a hierarchical generative adversarial network \cite{goodfellow2020generative}.  (5) \textbf{DRNet} \cite{schwab2020learning}, a multi-head deep model stratified according to treatment. (6) \textbf{VCNet} \cite{nie2021vcnet}, a varying coefficient neural network with functional targeted regularization. 

For all experiments, we perform 30 replications to report the mean integrated square error (MISE) and the standard deviations (std) of HDRC estimation: $\mathrm{MISE}=s^{-1}\sum_{i=1}^s$ $ \int_{{a}}^{{b}}(h(t, X_i)-\hat{h}({t},{X}_i))^2 d {t},$ where $s$ is the test sample size and $[{a}, {b}]$ is the sampling interval of treatment values.

\begin{table*}[t]

  \centering
  \resizebox{\linewidth}{!}{
    \begin{tabular}{lccccc|ccccccccc}
      \toprule
      Method & Data-1 & Data-2 & Data-5 & Data-10 & IHDP-1 & News-2 & News-4 & News-8 & News-16 \\
      \midrule
      GPS  & 57.7 ± 18 & 57.8 ± 14 & 57.4 ± 15 & 78.0 ± 19 & 0.98 ± 0.4 & 84.3 ± 4.6 & 83.8 ± 4.5 & 87.8 ± 3.8 & 89.0 ± 4.2 \\
      CBGPS  & 57.8 ± 18 & 57.8 ± 14 & 57.3 ± 15 & 70.5 ± 19 & 1.03 ± 0.4 & 84.1 ± 4.4 & 83.6 ± 4.6 & 85.5 ± 5.0 & 86.7 ± 4.5 \\
      CF  & \underline{1.83 ± 0.6} & \underline{2.50 ± 0.7} & 5.16 ± 0.9 & 14.9 ± 2.4 & 0.79 ± 0.3 & 26.8 ± 15 & 26.9 ± 11 & 47.9 ± 21 & 82.7 ± 81 \\
      DRNet  & 2.35 ± 0.7 & 3.49 ± 1.4 & 6.39 ± 2.1 & 18.5 ± 4.7 & 1.29 ± 0.4 & 18.0 ± 8.9 & 18.6 ± 10 & 33.3 ± 65 & 26.1 ± 10 \\
      SCIGAN & 15.0 ± 13 & 26.1 ± 13 & 43.6 ± 15 & 59.6 ± 26 & 0.65 ± 0.3 & 233 ± 218 & 163 ± 151 & 254 ± 365 & 200 ± 248 \\
      VCNet & 5.79 ± 4.8 & 6.41 ± 4.7 & 13.7 ± 5.7 & 28.2 ± 7.1 & 1.28 ± 0.7 & 11.3 ± 6.0 & 9.80 ± 3.3 & 26.5 ± 51 & 25.3 ± 31 \\
      \midrule
      CRNet & \textbf{1.69 ± 0.5} & \textbf{2.07 ± 0.8} & \textbf{3.05 ± 0.7} & \textbf{7.55 ± 2.6} & \textbf{0.22 ± 0.1} & \textbf{3.21 ± 1.4} & \textbf{5.19 ± 2.3} & \textbf{8.35 ± 5.0} & \textbf{9.18 ± 2.9} \\
      w/o BR & 2.04 ± 0.5 & {2.56 ± 1.0} & \underline{4.76 ± 1.2} & \underline{9.69 ± 5.1} & \underline{0.63 ± 0.4} & \underline{6.03 ± 4.6} & \underline{5.60 ± 3.4} & \underline{9.88 ± 5.9} & \underline{15.9 ± 23} \\
      w/o PR & 52.9 ± 16 & 53.9 ± 14 & 51.0 ± 13 & 55.2 ± 16 & 0.92 ± 0.4 & 35.3 ± 17 & 36.1 ± 17 & 36.1 ± 15 & 38.3 ± 14 \\
      \bottomrule
    \end{tabular}}
      \caption{Performance comparison (MISE $\pm$ std) and ablation studies on simulation Data-$q$-$p$, IHDP-$q$ and News-$q$.}
  \label{combined_results}
\end{table*}

\subsection{Results}

\paragraph{Performance Comparison.} We conduct simulation and semi-simulation experiments as shown in Table \ref{combined_results}, where bold indicates optimal performance, and underlined indicates suboptimal performance.
As the dimensionality of treatments increases, traditional statistical methods tend to fail, highlighting their limitations in handling complex, high-dimensional data. CRNet surpasses both DRNet and VCNet in performance, underscoring that relying solely on treatment-balanced representation or balancing representation can indeed lead to a loss in predictive capabilities. SCIGAN's poor performance in high-dimensional data reflects the instability inherent in generative adversarial networks while also emphasizing the necessity of balancing representation condition.
CRNet attains a state-of-the-art performance level in all conducted experiments. This demonstrates the effectiveness of the double balancing representation condition in enhancing both the constraint on unbiasedness and the outcome predictive capacity.

\paragraph{Ablation Studies.} To verity the performance of prognostic representation, we conduct the w/o balancing (BR) ablation study on CRNet with hyperparameter $\alpha=0$. To verify the performance of balancing representation, we conduct the w/o prognostic (PR) ablation study on CRNet, which applies a two-stage training strategy: Only loss $\ell^{CR}_{\Phi}$ is used in the first stage, and only $\ell^{MSE}$ loss is used in the second stage.
The results are shown in Table \ref{combined_results}. Although w/o balancing achieved good performance in most settings, its performance was still significantly degraded compared to CRNet. On the other hand, w/o prognostic performed poorly in all settings. 
This result aligns with our expectations since the model's predictive accuracy deteriorates when the prognostic representation condition is unsatisfied, and the sole reliance on prognostic representation proves biased in practice.

\begin{table}[tbp]
    \centering
    \scalebox{.92}{
        \begin{tabular}{l|ccccccc}
            \toprule
            $n*m$ & Data-1 & Data-10 & IHDP-1 & News-16 \\
            \midrule
            $m=0$  & 1.99 ± 0.6 & 11.0 ± 3.1 & 0.58 ± 0.3 & 10.2 ± 2.9 \\
            $m=1$ & 1.69 ± 0.5 & 7.55 ± 2.6 & \textbf{0.22 ± 0.1} & \textbf{9.18 ± 2.9} \\
            $m=2$ & \textbf{1.69 ± 0.5} & \textbf{7.06 ± 2.4} & 0.24 ± 0.1 & \underline{10.1 ± 4.0} \\
            $m=3$ & \underline{1.69 ± 0.5} & \underline{7.38 ± 3.2} & 0.23 ± 0.1 & 10.9 ± 5.0 \\
            $m=5$ & 1.71 ± 0.6 & 8.08 ± 3.1 & 0.23 ± 0.1 & 10.5 ± 4.1 \\
            $m=10$ & 1.70 ± 0.6 & 7.36 ± 2.7 & \underline{0.23 ± 0.1} & 11.5 ± 6.4 \\
            \bottomrule
        \end{tabular} 
    }
\caption{Performance comparison (MISE ± std) varying values $m$ of the number of negative sample augmentations.}\label{tab:neg_sample}
\end{table}

\paragraph{Hyperparameters Tuning.}

\begin{figure}[tbp]
  \centering
  \includegraphics[width=.95\linewidth]{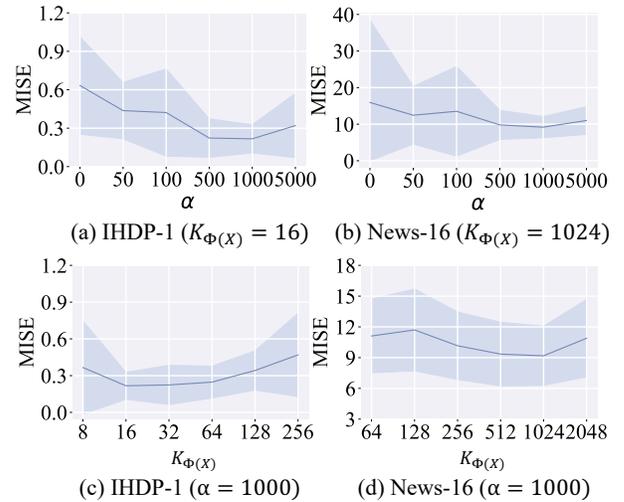}
  \caption{The sensitivity experiments (MISE $\pm$ SD) for the value of $\alpha$ and the dimension of double balancing representation $K_{\Phi({X})}$ on IHDP-1 and News-16 datasets.}
  \label{fig:hyperparameter}
\end{figure}

We conduct experiments to evaluate the impact of hyperparameters $\alpha$ in Eq. (\ref{finalloss}), the dimension of double balancing representation $K_{\Phi(X)}$, the augmentation of negative samples $m$ on the performance of CRNet. As Fig. \ref{fig:hyperparameter} shown, we found that a large $\alpha$ improves estimation performance. Nevertheless, when $\alpha$ is too large, it will be an obstacle to fitting the outcome. Moreover, we found that increasing the dimension $K_{\Phi(X)}$ does not lead to a substantial improvement in estimation performance, which implies that CRNet is not sensitive to the dimension. 

We further conduct experiments, as shown in Table \ref{tab:neg_sample}, by increasing the number of shuffle times  $m$ from 0 to 10. The results show a degradation in performance when $m=0$, highlighting that naive minimization of the partial distance measure can induce mode collapse. Conversely, increasing $m$ to 1 significantly improves performance, indicating the effectiveness of our designed negative sample constraint. This issue is particularly pronounced in high-dimensional datasets such as Data-10 and News-16. The results also show that our best results from the main text (Tables \ref{combined_results}) can be further improved by increasing the number of $m$. To enhance training efficiency, this paper defaults to $m=1$.

\section{Conclusion}

For estimating heterogeneous dose-response curves, we propose a neural network called CRNet. With no break of the continuity of treatments, this network discriminates between positive samples $\mathbf{X}$ and negative samples $\mathbf{X}'$ by a partial distance measure applied to double balancing representation. 
By employing this network, we enforce unbiasedness in our estimates and enable us to capture the prognostication present among individuals. 
Besides, this study has some potential limitations. We assume unconfoundedness and although we make efforts to control for potential confounding factors, there remains a possibility that unmeasured or unknown confounders may influence the results. Additionally, different conditional dependence measures may lead to different conclusions about the performance of our method.

\appendix

\section{Acknowledgments}
This work was supported in part by Young Elite Scientists Sponsorship Program by CAST (2021QNRC001), National Natural Science Foundation of China (No. 62376243, U20A20387), the StarryNight Science Fund of Zhejiang University Shanghai Institute for Advanced Study (SN-ZJU-SIAS-0010), Project by Shanghai AI Laboratory (P22KS00111) and Program of Zhejiang Province Science and Technology (2022C01044). Anpeng Wu's research was supported by the China Scholarship Council.

\bibliography{aaai24}

\begin{thebibliography}{53}
\providecommand{\natexlab}[1]{#1}

\bibitem[{Arbour, Dimmery, and Sondhi(2021)}]{arbour2021permutation}
Arbour, D.; Dimmery, D.; and Sondhi, A. 2021.
\newblock Permutation weighting.
\newblock In \emph{International Conference on Machine Learning}, 331--341. PMLR.

\bibitem[{Assaad et~al.(2021)Assaad, Zeng, Tao, Datta, Mehta, Henao, Li, and Carin}]{assaad2021counterfactual}
Assaad, S.; Zeng, S.; Tao, C.; Datta, S.; Mehta, N.; Henao, R.; Li, F.; and Carin, L. 2021.
\newblock Counterfactual representation learning with balancing weights.
\newblock In \emph{International Conference on Artificial Intelligence and Statistics}, 1972--1980. PMLR.

\bibitem[{Behrmann et~al.(2019)Behrmann, Grathwohl, Chen, Duvenaud, and Jacobsen}]{behrmann2019invertible}
Behrmann, J.; Grathwohl, W.; Chen, R.~T.; Duvenaud, D.; and Jacobsen, J.-H. 2019.
\newblock Invertible residual networks.
\newblock In \emph{International Conference on Machine Learning}, 573--582. PMLR.

\bibitem[{Bica, Jordon, and van~der Schaar(2020)}]{bica2020estimating}
Bica, I.; Jordon, J.; and van~der Schaar, M. 2020.
\newblock Estimating the effects of continuous-valued interventions using generative adversarial networks.
\newblock \emph{Advances in Neural Information Processing Systems}, 33: 16434--16445.

\bibitem[{Chen et~al.(2020)Chen, Kornblith, Norouzi, and Hinton}]{chen2020simple}
Chen, T.; Kornblith, S.; Norouzi, M.; and Hinton, G. 2020.
\newblock A simple framework for contrastive learning of visual representations.
\newblock In \emph{International Conference on Machine Learning}, 1597--1607. PMLR.

\bibitem[{Chen and He(2021)}]{chen2021exploring}
Chen, X.; and He, K. 2021.
\newblock Exploring simple siamese representation learning.
\newblock In \emph{Proceedings of the IEEE/CVF Conference on Computer Vision and Pattern Recognition}, 15750--15758.

\bibitem[{Cheng et~al.(2020)Cheng, Hao, Dai, Liu, Gan, and Carin}]{cheng2020club}
Cheng, P.; Hao, W.; Dai, S.; Liu, J.; Gan, Z.; and Carin, L. 2020.
\newblock Club: A contrastive log-ratio upper bound of mutual information.
\newblock In \emph{International Conference on Machine Learning}, 1779--1788. PMLR.

\bibitem[{Fong, Hazlett, and Imai(2018)}]{fong2018covariate}
Fong, C.; Hazlett, C.; and Imai, K. 2018.
\newblock Covariate balancing propensity score for a continuous treatment: Application to the efficacy of political advertisements.
\newblock \emph{The Annals of Applied Statistics}, 12(1): 156--177.

\bibitem[{Gan et~al.(2023)Gan, Li, Kuang, Zhang, Wang, Luu, Yang, and Wu}]{gan-etal-2023-exploiting}
Gan, L.; Li, B.; Kuang, K.; Zhang, Y.; Wang, L.; Luu, A.; Yang, Y.; and Wu, F. 2023.
\newblock Exploiting Contrastive Learning and Numerical Evidence for Confusing Legal Judgment Prediction.
\newblock In \emph{Findings of the Association for Computational Linguistics: EMNLP 2023}, 12174--12185.

\bibitem[{Goodfellow et~al.(2020)Goodfellow, Pouget-Abadie, Mirza, Xu, Warde-Farley, Ozair, Courville, and Bengio}]{goodfellow2020generative}
Goodfellow, I.; Pouget-Abadie, J.; Mirza, M.; Xu, B.; Warde-Farley, D.; Ozair, S.; Courville, A.; and Bengio, Y. 2020.
\newblock Generative adversarial networks.
\newblock \emph{Communications of the ACM}, 63(11): 139--144.

\bibitem[{Grill et~al.(2020)Grill, Strub, Altch{\'e}, Tallec, Richemond, Buchatskaya, Doersch, Avila~Pires, Guo, Gheshlaghi~Azar et~al.}]{grill2020bootstrap}
Grill, J.-B.; Strub, F.; Altch{\'e}, F.; Tallec, C.; Richemond, P.; Buchatskaya, E.; Doersch, C.; Avila~Pires, B.; Guo, Z.; Gheshlaghi~Azar, M.; et~al. 2020.
\newblock Bootstrap your own latent-a new approach to self-supervised learning.
\newblock \emph{Advances in neural information processing systems}, 33: 21271--21284.

\bibitem[{Hahn(1998)}]{hahn1998role}
Hahn, J. 1998.
\newblock On the role of the propensity score in efficient semiparametric estimation of average treatment effects.
\newblock \emph{Econometrica}, 315--331.

\bibitem[{Hainmueller(2012)}]{hainmueller2012entropy}
Hainmueller, J. 2012.
\newblock Entropy balancing for causal effects: A multivariate reweighting method to produce balanced samples in observational studies.
\newblock \emph{Political analysis}, 20(1): 25--46.

\bibitem[{Hansen(2008)}]{hansen2008prognostic}
Hansen, B.~B. 2008.
\newblock The prognostic analogue of the propensity score.
\newblock \emph{Biometrika}, 95(2): 481--488.

\bibitem[{He et~al.(2020)He, Fan, Wu, Xie, and Girshick}]{he2020momentum}
He, K.; Fan, H.; Wu, Y.; Xie, S.; and Girshick, R. 2020.
\newblock Momentum contrast for unsupervised visual representation learning.
\newblock In \emph{Proceedings of the IEEE/CVF conference on computer vision and pattern recognition}, 9729--9738.

\bibitem[{Hirano and Imbens(2004)}]{hirano2004propensity}
Hirano, K.; and Imbens, G.~W. 2004.
\newblock The propensity score with continuous treatments.
\newblock \emph{Applied Bayesian modeling and causal inference from incomplete-data perspectives}, 226164: 73--84.

\bibitem[{Holland(1986)}]{holland1986statistics}
Holland, P.~W. 1986.
\newblock Statistics and causal inference.
\newblock \emph{Journal of the American statistical Association}, 81(396): 945--960.

\bibitem[{Hu, Follmann, and Wang(2014)}]{hu2014estimation}
Hu, Z.; Follmann, D.~A.; and Wang, N. 2014.
\newblock Estimation of mean response via the effective balancing score.
\newblock \emph{Biometrika}, 101(3): 613--624.

\bibitem[{Huang and Chan(2017)}]{huang2017joint}
Huang, M.-Y.; and Chan, K. C.~G. 2017.
\newblock Joint sufficient dimension reduction and estimation of conditional and average treatment effects.
\newblock \emph{Biometrika}, 104(3): 583--596.

\bibitem[{Huang, Yi, and Zhao(2021)}]{huang2021towards}
Huang, W.; Yi, M.; and Zhao, X. 2021.
\newblock Towards the generalization of contrastive self-supervised learning.
\newblock \emph{arXiv preprint arXiv:2111.00743}.

\bibitem[{Imai and Ratkovic(2014)}]{imai2014covariate}
Imai, K.; and Ratkovic, M. 2014.
\newblock Covariate balancing propensity score.
\newblock \emph{Journal of the Royal Statistical Society: Series B (Statistical Methodology)}, 76(1): 243--263.

\bibitem[{Imai and Van~Dyk(2004)}]{imai2004causal}
Imai, K.; and Van~Dyk, D.~A. 2004.
\newblock Causal inference with general treatment regimes: Generalizing the propensity score.
\newblock \emph{Journal of the American Statistical Association}, 99(467): 854--866.

\bibitem[{Imbens(2000)}]{imbens2000role}
Imbens, G.~W. 2000.
\newblock The role of the propensity score in estimating dose-response functions.
\newblock \emph{Biometrika}, 87(3): 706--710.

\bibitem[{Jing et~al.(2021)Jing, Vincent, LeCun, and Tian}]{jing2021understanding}
Jing, L.; Vincent, P.; LeCun, Y.; and Tian, Y. 2021.
\newblock Understanding Dimensional Collapse in Contrastive Self-supervised Learning.
\newblock In \emph{International Conference on Learning Representations}.

\bibitem[{Kallus(2020)}]{kallus2020deepmatch}
Kallus, N. 2020.
\newblock Deepmatch: Balancing deep covariate representations for causal inference using adversarial training.
\newblock In \emph{International Conference on Machine Learning}, 5067--5077. PMLR.

\bibitem[{Kennedy et~al.(2017)Kennedy, Ma, McHugh, and Small}]{kennedy2017non}
Kennedy, E.~H.; Ma, Z.; McHugh, M.~D.; and Small, D.~S. 2017.
\newblock Non-parametric methods for doubly robust estimation of continuous treatment effects.
\newblock \emph{Journal of the Royal Statistical Society. Series B (Statistical Methodology)}, 79(4): 1229--1245.

\bibitem[{LeCun et~al.(2006)LeCun, Chopra, Hadsell, Ranzato, and Huang}]{lecun2006tutorial}
LeCun, Y.; Chopra, S.; Hadsell, R.; Ranzato, M.; and Huang, F. 2006.
\newblock A tutorial on energy-based learning.
\newblock \emph{Predicting structured data}, 1(0).

\bibitem[{Lee and Lee(2022)}]{lee2022review}
Lee, M.-J.; and Lee, S. 2022.
\newblock Review and comparison of treatment effect estimators using propensity and prognostic scores.
\newblock \emph{The international journal of biostatistics}, 18(2): 357--380.

\bibitem[{Li et~al.(2023{\natexlab{a}})Li, Xiao, Zheng, Wu, and Cui}]{li2023propensity}
Li, H.; Xiao, Y.; Zheng, C.; Wu, P.; and Cui, P. 2023{\natexlab{a}}.
\newblock Propensity matters: Measuring and enhancing balancing for recommendation.
\newblock In \emph{International Conference on Machine Learning}, 20182--20194. PMLR.

\bibitem[{Li et~al.(2023{\natexlab{b}})Li, Zheng, Cao, Geng, Liu, and Wu}]{li2023trustworthy}
Li, H.; Zheng, C.; Cao, Y.; Geng, Z.; Liu, Y.; and Wu, P. 2023{\natexlab{b}}.
\newblock Trustworthy policy learning under the counterfactual no-harm criterion.
\newblock In \emph{International Conference on Machine Learning}, 20575--20598. PMLR.

\bibitem[{Li et~al.(2023{\natexlab{c}})Li, Zheng, Wu, Kuang, Liu, and Cui}]{li2023should}
Li, H.; Zheng, C.; Wu, P.; Kuang, K.; Liu, Y.; and Cui, P. 2023{\natexlab{c}}.
\newblock Who should be given incentives? counterfactual optimal treatment regimes learning for recommendation.
\newblock In \emph{Proceedings of the 29th ACM SIGKDD Conference on Knowledge Discovery and Data Mining}, 1235--1247.

\bibitem[{Mealli, Pacini, and Rubin(2011)}]{mealli2011statistical}
Mealli, F.; Pacini, B.; and Rubin, D.~B. 2011.
\newblock Statistical inference for causal effects.
\newblock \emph{Modern analysis of customer surveys: With applications using R}, 171--192.

\bibitem[{Nie et~al.(2021)Nie, Ye, Liu, and Nicolae}]{nie2021vcnet}
Nie, L.; Ye, M.; Liu, Q.; and Nicolae, D. 2021.
\newblock Vcnet and functional targeted regularization for learning causal effects of continuous treatments.
\newblock In \emph{International Conference on Learning Representations}.

\bibitem[{Pearl(2009)}]{pearl2009causality}
Pearl, J. 2009.
\newblock \emph{Causality}.
\newblock Cambridge university press.

\bibitem[{Raita et~al.(2021)Raita, Camargo~Jr, Liang, and Hasegawa}]{raita2021leveraging}
Raita, Y.; Camargo~Jr, C.~A.; Liang, L.; and Hasegawa, K. 2021.
\newblock Leveraging “big data” in respiratory medicine--data science, causal inference, and precision medicine.
\newblock \emph{Expert Review of Respiratory Medicine}, 15(6): 717--721.

\bibitem[{Rosenbaum and Rubin(1983)}]{rosenbaum1983central}
Rosenbaum, P.~R.; and Rubin, D.~B. 1983.
\newblock The central role of the propensity score in observational studies for causal effects.
\newblock \emph{Biometrika}, 70(1): 41--55.

\bibitem[{Rubin(1974)}]{rubin1974estimating}
Rubin, D.~B. 1974.
\newblock Estimating causal effects of treatments in randomized and nonrandomized studies.
\newblock \emph{Journal of educational Psychology}, 66(5): 688.

\bibitem[{Schwab et~al.(2020)Schwab, Linhardt, Bauer, Buhmann, and Karlen}]{schwab2020learning}
Schwab, P.; Linhardt, L.; Bauer, S.; Buhmann, J.~M.; and Karlen, W. 2020.
\newblock Learning counterfactual representations for estimating individual dose-response curves.
\newblock In \emph{Proceedings of the AAAI Conference on Artificial Intelligence}, volume~34, 5612--5619.

\bibitem[{Shalit, Johansson, and Sontag(2017)}]{shalit2017estimating}
Shalit, U.; Johansson, F.~D.; and Sontag, D. 2017.
\newblock Estimating individual treatment effect: generalization bounds and algorithms.
\newblock In \emph{International Conference on Machine Learning}, 3076--3085. PMLR.

\bibitem[{Shi, Miao, and Tchetgen(2020)}]{shi2020selective}
Shi, X.; Miao, W.; and Tchetgen, E.~T. 2020.
\newblock A selective review of negative control methods in epidemiology.
\newblock \emph{Current epidemiology reports}, 7(4): 190--202.

\bibitem[{Stuart, Lee, and Leacy(2013)}]{stuart2013prognostic}
Stuart, E.~A.; Lee, B.~K.; and Leacy, F.~P. 2013.
\newblock Prognostic score--based balance measures can be a useful diagnostic for propensity score methods in comparative effectiveness research.
\newblock \emph{Journal of clinical epidemiology}, 66(8): S84--S90.

\bibitem[{Sz{\'e}kely and Rizzo(2014)}]{szekely2014partial}
Sz{\'e}kely, G.~J.; and Rizzo, M.~L. 2014.
\newblock Partial distance correlation with methods for dissimilarities.
\newblock \emph{The Annals of Statistics}, 42(6): 2382--2412.

\bibitem[{Vegetabile et~al.(2021)Vegetabile, Griffin, Coffman, Cefalu, Robbins, and McCaffrey}]{vegetabile2021nonparametric}
Vegetabile, B.~G.; Griffin, B.~A.; Coffman, D.~L.; Cefalu, M.; Robbins, M.~W.; and McCaffrey, D.~F. 2021.
\newblock Nonparametric estimation of population average dose-response curves using entropy balancing weights for continuous exposures.
\newblock \emph{Health Services and Outcomes Research Methodology}, 21(1): 69--110.

\bibitem[{Wager and Athey(2018)}]{wager2018estimation}
Wager, S.; and Athey, S. 2018.
\newblock Estimation and inference of heterogeneous treatment effects using random forests.
\newblock \emph{Journal of the American Statistical Association}, 113(523): 1228--1242.

\bibitem[{Wang et~al.(2023)Wang, Chen, Fan, Li, Liu, Liu, Dai, Wang, Dong, and Tang}]{wang2023optimal}
Wang, H.; Chen, Z.; Fan, J.; Li, H.; Liu, T.; Liu, W.; Dai, Q.; Wang, Y.; Dong, Z.; and Tang, R. 2023.
\newblock Optimal transport for treatment effect estimation.
\newblock \emph{Advances in Neural Information Processing Systems}.

\bibitem[{Wang and Isola(2020)}]{wang2020understanding}
Wang, T.; and Isola, P. 2020.
\newblock Understanding contrastive representation learning through alignment and uniformity on the hypersphere.
\newblock In \emph{International Conference on Machine Learning}, 9929--9939. PMLR.

\bibitem[{Wu et~al.(2022{\natexlab{a}})Wu, Kuang, Li, and Wu}]{wu2022instrumental}
Wu, A.; Kuang, K.; Li, B.; and Wu, F. 2022{\natexlab{a}}.
\newblock Instrumental variable regression with confounder balancing.
\newblock In \emph{International Conference on Machine Learning}, 24056--24075. PMLR.

\bibitem[{Wu et~al.(2023)Wu, Kuang, Xiong, Li, and Wu}]{wu2023stable}
Wu, A.; Kuang, K.; Xiong, R.; Li, B.; and Wu, F. 2023.
\newblock Stable estimation of heterogeneous treatment effects.
\newblock In \emph{International Conference on Machine Learning}, 37496--37510. PMLR.

\bibitem[{Wu et~al.(2022{\natexlab{b}})Wu, Yuan, Kuang, Li, Wu, Zhu, Zhuang, and Wu}]{wu2022learning}
Wu, A.; Yuan, J.; Kuang, K.; Li, B.; Wu, R.; Zhu, Q.; Zhuang, Y.; and Wu, F. 2022{\natexlab{b}}.
\newblock Learning decomposed representations for treatment effect estimation.
\newblock \emph{IEEE Transactions on Knowledge and Data Engineering}, 35(5): 4989--5001.

\bibitem[{Wu et~al.(2018)Wu, Xiong, Yu, and Lin}]{wu2018unsupervised}
Wu, Z.; Xiong, Y.; Yu, S.~X.; and Lin, D. 2018.
\newblock Unsupervised feature learning via non-parametric instance discrimination.
\newblock In \emph{Proceedings of the IEEE conference on computer vision and pattern recognition}, 3733--3742.

\bibitem[{Yao et~al.(2022)Yao, Zhao, Zhang, Zhu, Zhu, Zhang, and He}]{DBLP:conf/www/YaoZZZZZH22}
Yao, D.; Zhao, Z.; Zhang, S.; Zhu, J.; Zhu, Y.; Zhang, R.; and He, X. 2022.
\newblock Contrastive Learning with Positive-Negative Frame Mask for Music Representation.
\newblock In \emph{{WWW} '22: The {ACM} Web Conference 2022, Virtual Event, Lyon, France, April 25 - 29, 2022}, 2906--2915. {ACM}.

\bibitem[{Zhang et~al.(2022)Zhang, Huang, Li, and Wang}]{zhang2022tree}
Zhang, M.; Huang, S.; Li, W.; and Wang, D. 2022.
\newblock Tree structure-aware few-shot image classification via hierarchical aggregation.
\newblock In \emph{European Conference on Computer Vision}, 453--470. Springer.

\bibitem[{Zhang et~al.(2021)Zhang, Yao, Zhao, Chua, and Wu}]{DBLP:conf/sigir/ZhangYZC021}
Zhang, S.; Yao, D.; Zhao, Z.; Chua, T.; and Wu, F. 2021.
\newblock CauseRec: Counterfactual User Sequence Synthesis for Sequential Recommendation.
\newblock In \emph{{SIGIR} '21: The 44th International {ACM} {SIGIR} Conference on Research and Development in Information Retrieval, Virtual Event, Canada, July 11-15, 2021}, 367--377. {ACM}.

\end{thebibliography}

\end{document}